\documentclass[smallextended]{svjour3}       
\usepackage{graphicx}
\usepackage{siunitx}
\usepackage{cite}
\usepackage[FIGTOPCAP,raggedright]{subfigure}

\def\makeheadbox{\relax}
\begin{document}

\title{Fine-grained Apparel Classification and Retrieval without rich annotations
}


\author{Aniket Bhatnagar         \and
        Sanchit Aggarwal 
}

\institute{Aniket Bhatnagar \at
              Squadrun Solutions Private Limited \\
              Mobile.: +91 9999105171 \\
              \email{aniket@squadplatform.com}           
           \and
           Sanchit Aggarwal \at
              Squadrun Solutions Private Limited \\
              \email{sanchit@squadplatform.com}
}
\maketitle

\begin{abstract}
The ability to correctly classify and retrieve apparel images has a variety of applications
important to e-commerce, online advertising, internet search, and visual surveillance industry.
In this work, we propose a robust framework for fine-grained apparel classification,
in-shop and cross-domain retrieval which eliminates the requirement of rich annotations like bounding boxes
and human-joints or clothing landmarks, and training of bounding box/ key-landmark detector for the same.
Factors such as subtle appearance differences, variations in human poses, different shooting angles,
apparel deformations, and self-occlusion add to the challenges in classification and retrieval of apparel items.
Cross-domain retrieval is even harder due to the presence of large variation between online shopping images,
usually taken in ideal lighting, pose, positive angle and clean background as compared with street photos
captured by users in complicated conditions with poor lighting, cluttered scenes, and complex background.
Our framework utilizes compact bilinear CNN~\cite{gao2016compact} with tensor sketch algorithm to generate embeddings that
capture local pairwise feature interactions in a translationally invariant manner. For apparel classification,
we pass the obtained feature embeddings through a softmax classifier, while, the in-shop and cross-domain
retrieval pipelines use a triplet-loss based optimization approach and deploy three compact BCNNs,
with a ranking loss such that squared Euclidean distance between embeddings measures the dissimilarity between the images.
Unlike previous settings that relied on bounding box, key clothing landmarks or human joint detectors to assist the final
deep classifier, proposed framework can be trained directly on the provided category labels or generated triplets for
triplet loss optimization. Lastly, Experimental results on the DeepFashion ~\cite{liu2016deepfashion} fine-grained
categorization, and in-shop and consumer-to-shop retrieval datasets
provide a comparative analysis with previous work performed in the domain.

\end{abstract}

\section{Introduction}
\label{intro}

Many methods have been proposed on the subject of apparel recognition~\cite{kalantidis2013getting, liu2016deepfashion, yang2011real, bossard2012apparel}, clothes parsing~\cite{yamaguchi2012parsing, yang2014clothing, gallagher2008clothing, simo2014high, hasan2010segmentation, jammalamadaka2013parsing, wang2011blocks, yamaguchi2013paper}, apparel attribute detection and description~\cite{dong2017multi, chen2012describing, liuYLWTeccv16FashionLandmark, chen2015deep, kiapour2014hipster, simo2015neuroaesthetics}, clothing item retrieval and recommendation~\cite{jagadeesh2014large, wang2014learning, huang2015cross, liu2012street, liu2012hi, di2013style, hadi2015buy, wang2016matching, lin2015rapid, liang2016clothes} due to its tremendous impact on various industries.

Online retail stores itself have huge opportunities ranging from a rich user discovery experience to quality control operation such as product identification, tagging, moderation, enrichment, and contextual advertisement. Consequentially, algorithms for automatic categorization and retrieval of visually similar fashion products would have significant benefits.

However, clothes categorization and retrieval is still an open problem, especially due to a large number of fine-grained categories with very subtle visual differences in style, texture, and cutting, compared with other common object categories. It is even more challenging to classify apparel images due to factors such as appearance, variations of human poses, and different shooting angles.

Another challenge is the subjectivity of apparels to deformations and self-occlusion. Moreover, apparel retrieval is often confronted with difficulties due to large variations between online shopping images compared with selfies. Usually, online shopping images have ideal lighting, pose, clean backgrounds and are captured from a positive angle, while street photos captured by users have poor lighting, cluttered scenes, and complex background.

Last few years have seen an emergence in the domain of fine-grained categorization ~\cite{lin2015bilinear, branson2014bird, zhang2014part, wang2014learning}. Since, discriminating parts for the apparel categories tend to become subtle differences in shapes, styles and textures, categorization or identifying different attributes of a clothing item like sleeve length or pattern can be posed as a fine-grained classification problem.

Fine-grained classification compared to general purpose visual categorization problems, focuses on the characteristic challenge of making subtle distinctions despite high intra-class variance due to factors such as pose, viewpoint or location of the object. A common approach to fine-grained classification problem is to first localize various parts of the object and model the appearance conditioned on detected locations.

Parts are often defined manually and a part detector is also learned as part of the overall pipeline. Branson et al.~\cite{branson2014bird}, approached the problem for fine-grained visual categorization of bird species, by estimating objects pose and computing features by deploying deep convolutional nets to image patches that are located and normalized by the pose. Zhang et al.~\cite{zhang2014part} leveraged deep convolutional features computed on bottom-up region proposals. These methods generally increase the cost of labeling as annotating tags is easier than annotating bounding box coordinates for part detectors.

Recently, an easier method which does not require bounding boxes or landmark locations for fine-grained classification namely Bilinear Convolutional Neural Nets (BCNNs) have been introduced. Lin et al.~\cite{lin2015bilinear} proposed a framework which utilizes a layer of bilinear pooling, just before the last fully connected layer which helps achieve remarkable performance on fine-grained classification datasets.

Bilinear pooling collects second order statistics of local features over the whole image and then does a global pool operation across each channel. The second order statistics capture pairwise correlations between the feature channels and global pooling introduces invariance to deformations. However, the representational power of bilinear features comes with very high dimensional feature maps. To reduce the model size and end-to-end optimization of the visual recognition system, Gao et al.~\cite{gao2016compact}, proposed using Compact Bilinear CNN’s with TensorSketch or Random Maclaurin algorithm. Gao used a kernelized representation to exhibit that bilinear descriptor compares each local descriptor in the first image with that in the second image and the comparison operator is a second order polynomial kernel. Thus proving that a compact version of the bilinear pooling is possible using any low dimensional approximation of the second order polynomial kernel. Further, compact bilinear CNNs exhibit near equal and at times better performance as compared to full bilinear CNN.

Thus, in this work, we propose an efficient and reliable framework based on compact bilinear CNN for fine-grained apparel classification, in-shop, and cross-domain retrieval, eliminating requirements for bounding box or key landmarks. To our information, this is the first attempt at solving Fashion products (apparel items) classification and retrieval without using bounding boxes or finding key landmarks and employing a compact bilinear CNN to do the same.

\section{Related Work}
\label{related work}

We now take a closer look at the more recent work on Apparel Classification and Retrieval~\cite{liu2016deepfashion, dong2017multi, huang2015cross, kalantidis2013getting, bossard2012apparel, wang2014learning,  liu2012street, liu2012hi, liang2016clothes, hadi2015buy}. These methods are quite efficient although based on strong requirement of manually labelled clothing landmarks~\cite{liu2016deepfashion}, or object detectors to predict bounding boxes~\cite{dong2017multi, huang2015cross, liang2016clothes, hadi2015buy}, estimating human pose~\cite{kalantidis2013getting, liu2012street} or body parts detector~\cite{bossard2012apparel, liu2012hi}.

Liu et al.~\cite{liu2016deepfashion} proposed FahionNet, which tries to simultaneously model local attribute level, general category level, and clothing image similarity level representation with the dependence on clothing attributes and landmarks. Apart from this, FashionNet also requires bounding box annotation around clothing item or around human model wearing the clothing apparel in the image for learning classifiers. Obtaining these massive attribute annotations along with clothing landmarks for apparel items is a tedious and costly task. It is not always possible for online marketplaces which  maintain huge catalogues of clothing items to create such hand-crafted annotated datasets.

Dong et al.~\cite{dong2017multi}, construct a deep model capable of recognizing fine-grained clothing attributes on images in the wild using multi-task curriculum transfer learning.  They collected a large clothing dataset and their meta-label as attributes from different online shopping web-sites. They learned a pre-processor to detect clothing images using Faster R-CCN and then employed an object detector which was trained on PASCAL VOC2007 followed by fine-tuning on an assembled bounding box annotated clothing dataset consisting of 8, 000 street/shop photos. Their model was then trained using obtained bounding boxes and rich annotations.

Hadi et al.~\cite{hadi2015buy} utilize an alexnet~\cite{krizhevsky2012imagenet} with activations from a fully connected layer FC6 to identify exact matching clothes from street to shop domain. They collected street and shop photos, and obtained bounding box annotations using Mechanical Turk service.  Huang et al.~\cite{huang2015cross}, proposed a dual attribute-aware ranking network, which optimizes attribute classification loss and image triplet quantization loss together for cross-domain image retrieval. They utilize two-stream CNN for handling in- shop and street images respectively with the dependence on bounding boxes generated using Faster RCNN~\cite{ren2015faster}. They curated 381,975 online-offline image pairs
of different categories from the customer review pages. Then, manually pruned the noisy labels, merged similar labels based on human perception using crowd-source annotators and obtained fine grained clothing attributes  using image descriptors.

Bossard et al.~\cite{bossard2012apparel} introduce a recognition and classification pipelines which include building blocks like upper body detectors, feature channels and a multi class learner based on random forest. They crawled the dataset from web and defined 15 clothing classes and used a bounding box detector to label the images. Liang et al.~\cite{liang2016clothes}, developed an integrated system for cloth co-parsing using a multi-image graphical model. They constructed a clothes dataset which consisted of 2098 high resolution street fashion images and requested annotators to clean the dataset along with extraction from the text tags of images for semantic attribute labelling for their joint label formulation. The dependency on rich annotations for above fine-grained apparel classification and retrieval tasks requires huge time and cost to curate a dataset.

The main contribution of our work is a robust framework that can be used to categorize apparel images, and identify similar clothing items for both in-shop and cross-domain image retrieval problems, without any overhead to train a network to identify bounding boxes around clothing items or to identify landmark locations in an apparel image. For both image retrieval tasks [in-shop and cross domain], we avoided the daunting task of manual labelling of bounding boxes or landmarks locations and yet attain robustness with effective usage of compact bilinear CNN’s along with triplet loss approach. Rather than curating a new fashion dataset, we have shown the performance of our framework on existing benchmark datasets for fashion apparel categorization and image retrieval. In Section III, we give an overview of our framework and describe the dataset and features used for various models. Section IV describes the evaluation metrics followed by results and analysis of experiments in Section IV-B. We conclude the paper in Section V.

\section{Overview of our framework}
An overview of our frameworks for apparel categorization and retrieval pipelines is illustrated in Figure 1 and Figure 2 respectively. We aim to solve fashion categorization and retrieval of apparel items using two different frameworks based on compact bilinear CNN. This will help to solve the problem with greater performance and without any overhead to train a network to identify bounding boxes around clothing items or to identify landmark locations in a clothing image. The three problems which we tackled using the proposed framework are:

\begin{itemize}
\item \textbf{Fashion Apparel Categorization}: The goal here is to assign each apparel item a unique category amongst the fifty fine-grained yet mutually exclusive categories.

\begin{figure*}
  \includegraphics[width=1.0\textwidth]{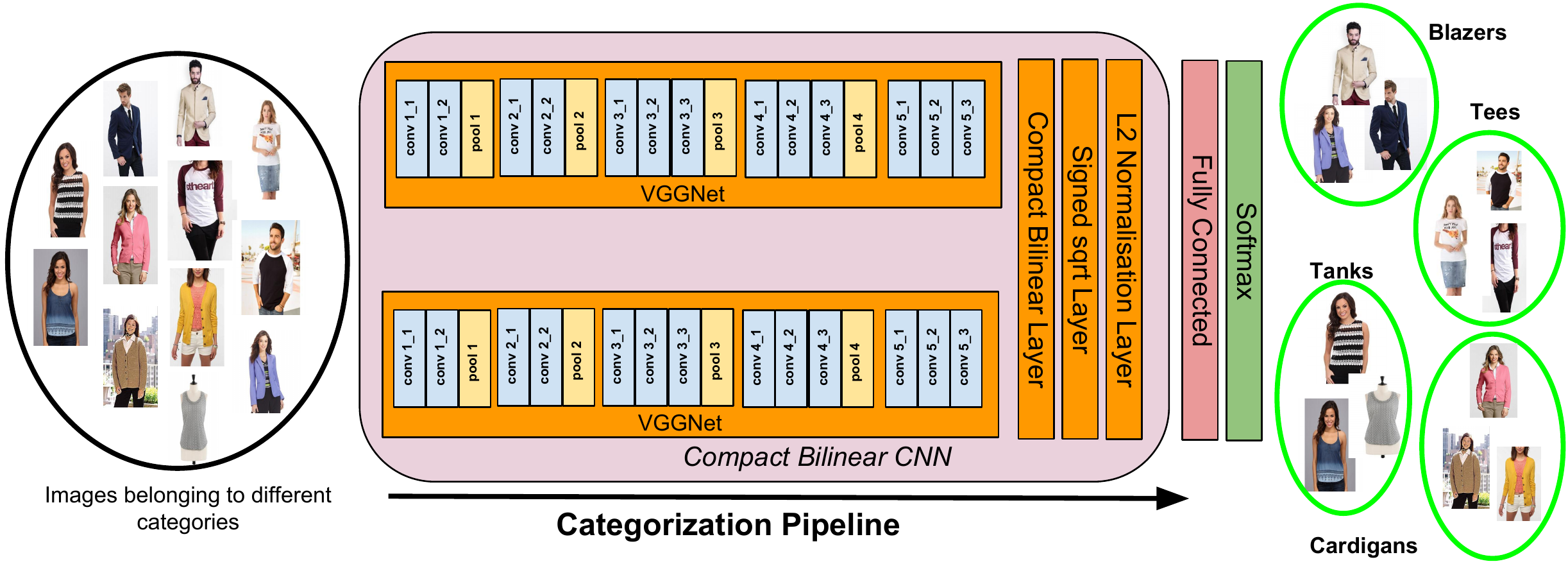}
\caption{Overview of categorisation pipeline. We utilize compact BCNNs~\cite{gao2016compact} with tensor sketch algorithm to generate embeddings that capture local pairwise feature interactions followed by a fully connected soft-max layer. (Best viewed in color)}
\label{fig:2}       
\end{figure*}

\item \textbf{Fashion Apparel in-shop image retrieval}: Given a clothing image, aim here is to identify whether two apparel images belong to the same item or not. It can be helpful when customers encounter a shop-image on a particular e-commerce website and want to know more details for the particular product or similar products on other e-commerce sites.

\item \textbf{Fashion Apparel cross-domain (street to shop) image retrieval}: Given a street image of clothing item in unconstrained domain i.e. consumer clicked photographs, the target here is to match it with its shop counterparts where images are taken in a constrained environment, i.e. by professional photographers under apt light and brightness levels. This can be useful for a person who wants to buy the same apparel as that seen on a friend or a celebrity picture.
\end{itemize}

\begin{figure*}
  \includegraphics[width=1.0\textwidth]{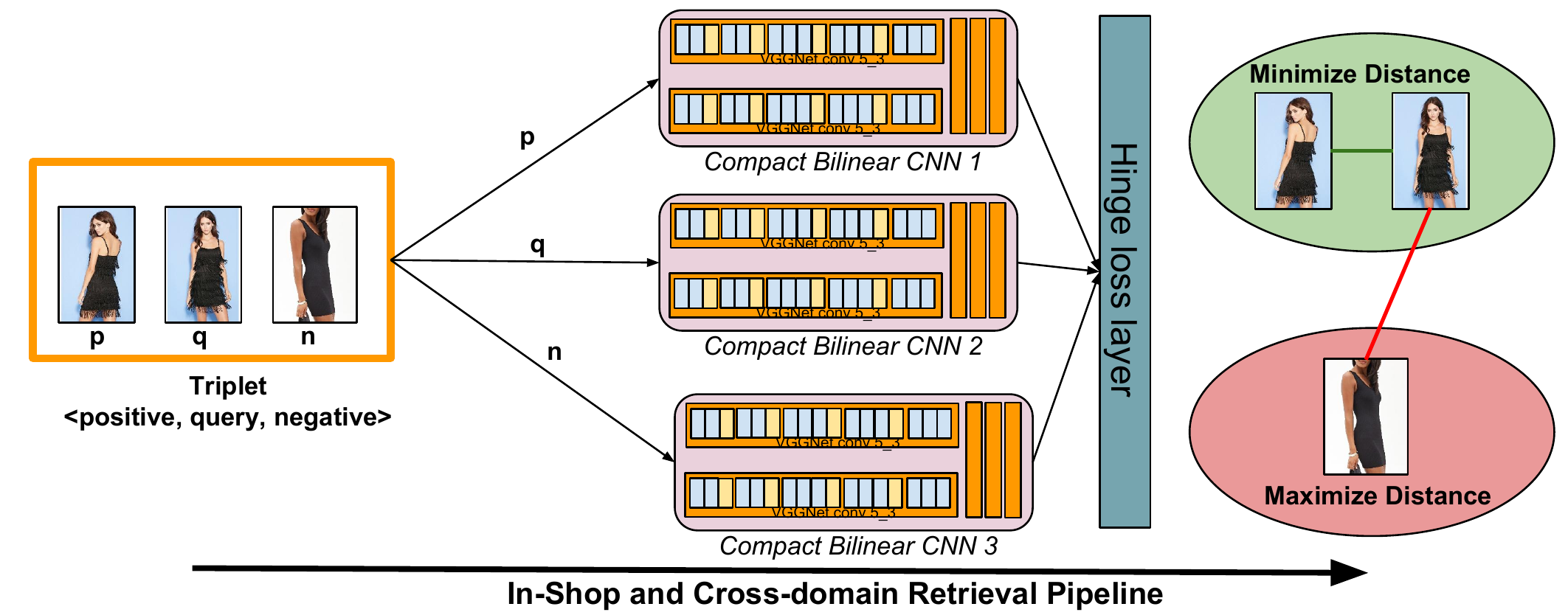}
\caption{Overview of in-shop and cross-domain retrieval pipeline. We use a triplet-based approach and deploy three compact BCNNs~\cite{gao2016compact} to generate feature embeddings, followed by hinge loss layer to minimze distance between similar images and maximise distance between dissimilar images}
\label{fig:2}       
\end{figure*}

Doing this we also highlight that compact BCNNs can be used to generate embeddings that capture the notion of visual similarity and later use these embeddings to find apparel item nearest to a query input fashion image.

We use the same architecture for compact BCNN as mentioned in~\cite{gao2016compact}. Further, we use Tensor Sketch algorithm instead of Random Maclaurin algorithm as it is faster and is more memory efficient. We deploy a symmetric compact bilinear CNN which uses VGG16 network~\cite{simonyan2014very} till the last convolution layer with Relu activation to obtain feature matrices. These feature matrices are followed by a compact bilinear layer which computes tensor-sketch projections of the embedding matrices and performs a sum pooling over all regions in the obtained projection matrices. The feature vectors obtained from VGGNet are of the order 1 x 512, and we set compact bilinear layer output dimensions [tensor sketch projection dimensions] to 8192 channels as it has been noted in~\cite{gao2016compact} that 8192-dimensional Tensor sketch features have the same performance as full-scale bilinear features. The resultant vector is passed through signed square root step followed by L2 normalization motivated from~\cite{perronnin2010improving}.

We used VGGNet instead of ResNet or Inception as the base model for the compact bilinear CNN pipeline as the primary purpose of Bilinear CNN or Compact bilinear CNN is to increase the representation power of CNN, without having to drastically increase the number of layers in the base network. Additionally, training a network like ResNet in place of VGG based compact BCNN would increase the number of weights to learn and the time it takes for end-to-end optimization of the model drastically.

Finally, the l2 normalized feature vector is pushed to fully connected softmax layer in case of categorization use case. The image retrieval model, for both in-shop and consumer-to-shop retrieval, uses a triplet-based approach with a ranking loss~\cite{wang2014learning} to learn embeddings such that squared Euclidean distance between embeddings measures the (dis)similarity between the images. The compact bilinear CNN based pipeline for each use case was implemented in Caffe and trained on NVIDIA K80 GPU.

\subsection{Pipeline and Model Training}

\subsubsection{Fine-Grained Categorization Model for Apparel Item}

We used the compact bilinear CNN with weights for layers before the compact bilinear layer initialized with an
imagenet~\cite{deng2009imagenet} pre-trained VGG weights. Similar to~\cite{chen2012describing}, we trained the compact
BCNN using the two-step procedure, where we initially train only the last fully connected layer at a large learning
rate of 1.0 and a small weight decay constant of \num{5e-6}  and then fine tune the entire model for several
iterations at a small learning rate of 0.001 and a larger weight decay constant of \num{5e-4}. Both training procedures were carried out using Momentum based SGD optimizer keeping momentum at 0.9. For image preprocessing, we resize all images to size 512 x 512 and cropped each image from the center for size 448 x 448. We then subtracted imagenet mean from each color channel for each image.

\subsubsection{Fine Grained Retrieval Model for In-Shop and Cross Domain Apparel Item Retrieval}
The image retrieval model was trained using the triplet of images provided as input to the triplet of compact bilinear
CNN models with shared weights. Each triplet of the form $< q, p, n >$ is pushed to their respective subnets.
Here q, p, and n represent the query image, matching image and any dissimilar image respectively.
We take the output from l2- normalization layer of each C-BCNN to represent embeddings for each image in the triplet.
The three subnets share the same weights in the entire training process. These embeddings $<q_{vec}, p_{vec}, n_{vec}>$
are then fed to a hinge loss function to optimize the network to be able to differentiate between similar and dissimilar
images. We use the following loss function:

\begin{equation}
               L = max(0, g + D(p_{vec}, q_{vec}) - D(q_{vec}, n_{vec}))
\end{equation}

where D(x, y) represents the squared Euclidean distance between two embedding vectors. Also note as embedding vectors are
L2 normalized the squared euclidean distance is equal to twice of cosine distance between the embedding vectors.
Besides this, the number of output dimensions for compact bilinear layer [or dimension of tensor-sketch projections]
remains the same as 8192. Thus each image embedding has a length of 8192. The hinge loss optimization helps
pull $q_{vec}$ and $p_{vec}$ closer while pushing the embedding vectors $q_{vec}$ and $n_{vec}$ farther.

We used triplet loss paradigm to train the model for concept of similarity between apparel item images, rather than using
siamese network ~\cite{bromley1994signature}, as triplet loss paradigm on each iteration trains the model to minimise
distance between similar inputs and maximise distance between dissimilar inputs simultaneously, while siamese network
would on a given iteration only look at a pair of images. Thus, triplet loss based trained models should be better able to
learn the context of "why" item "q" is closer to item "p", and not "n".

For this network, same image pre-processing was applied as for fine-grained categorization model. The model weights
for each of the subnets were initialized from the weights of the fine-tuned C-BCNN used for categorization. The
complete network was then trained at a small learning rate of 0.001, with a weight decay constant of \num{5e-4},
keeping g=1 in the triplet loss function. Here also momentum based SGD optimizer is used while training at a momentum
of 0.9.

\section{EXPERIMENTS AND EVALUATION}

\subsection{Dataset Used}
We use DeepFashion~\cite{liu2016deepfashion} fine-grained categorization benchmark, and in-shop clothing benchmark dataset to measure the performance of Compact BCNN with FashionNet~\cite{liu2016deepfashion}, DARN~\cite{huang2015cross} and WTBI~\cite{chen2012describing}. DeepFashion categorization data consists of nearly 200, 000 training images, and 40, 000 each validation and test images with 50 fine-grained clothing apparel categories, while the in-shop clothes retrieval benchmark consists of 52712 images with 7982 unique clothing items.

\subsubsection{Fashion Apparel Categorization}
In fine-grained category classification compact bilinear CNN outperforms FashionNet~\cite{liu2016deepfashion}, DARN~\cite{huang2015cross} and WTBI~\cite{chen2012describing} without using either bounding box annotation or clothing landmarks. Thus, performance boost with a much less expensive labeling technique. Table I specifies a quantitative comparison of our apparel categorization framework while Figure 3 displays the qualitative results of the pipeline.

This performance even in the absence of bounding box or landmark detector can be attributed to tensor sketch projections
which have the ability to represent pairwise correlations between the feature channels in a concise manner.
Thus, assisting the model to focus on the features which would help it in uniquely identifying the fine grained category
in relation to the object. Example, visualisations from Conv-5-3 layer of categorisation pipeline (Figure 4) using concept of
class activation mapping \cite{selvaraju2017grad} showcases that compact bilinear layer infuses the principle of
attention in the model without explicit use of an attention layer, and thus helping it centre attention on key-landmarks
without having to train the model for the same.

\begin{table}[t]
\renewcommand{\arraystretch}{1}
\caption{Top-3 and Top-5 Accuracy for Category Classification. The proposed method is significantly better than other state of the art teachniques. We achieved these result without the dependence on Part based models or rich annotations required for labelling datasets.}
\label{tab:Result1}
\centering
\begin{tabular}{|l|c|c|}
\hline \textbf{Method} & \textbf{Top-3}  & \textbf{Top-5}\\
\hline WTBI~\cite{chen2012describing} &     43.73   &   66.26\\
\hline DARN~\cite{huang2015cross} &    59.48 &    79.58\\
\hline FashionNet~\cite{liu2016deepfashion} &    82.58 &    90.17\\
\hline Categorisation Framework & \textbf{84.97} &    \textbf{91.69}\\
\hline
\end{tabular}
\end{table}

\begin{figure*}[!htb] 
\begin{center}
\subfigure[a)][]{\includegraphics[width=.375in,height=.375in]{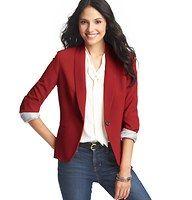}\label{Image}}
\subfigure[b)][]{\includegraphics[width=.375in,height=.375in]{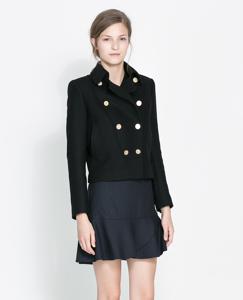}\label{Image}}
\subfigure[c)][]{\includegraphics[width=.375in,height=.375in]{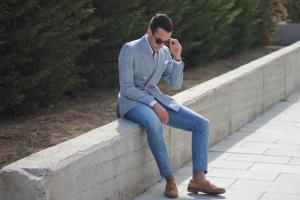}\label{Image}}
\subfigure[d)][]{\includegraphics[width=.375in,height=.375in]{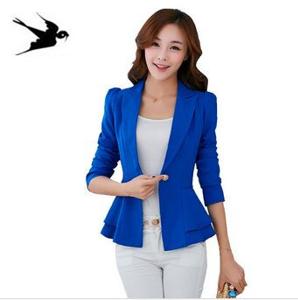}\label{Image}}
\subfigure[e)][]{\includegraphics[width=.375in,height=.375in]{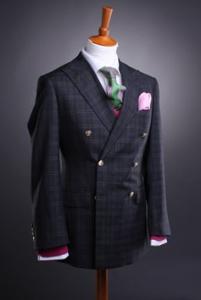}\label{Image}}
\subfigure[f)][]{\includegraphics[width=.375in,height=.375in]{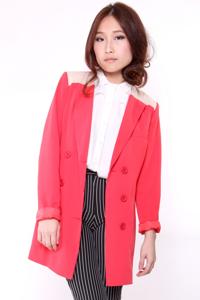}\label{Image}}
\subfigure[g)][]{\includegraphics[width=.375in,height=.375in]{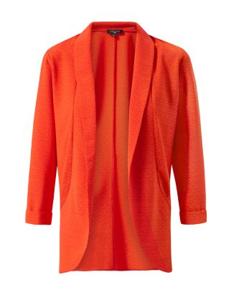}\label{Image}}
\subfigure[h)][]{\includegraphics[width=.375in,height=.375in]{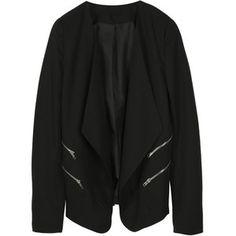}\label{Image}}
\subfigure[i)][]{\includegraphics[width=.375in,height=.375in]{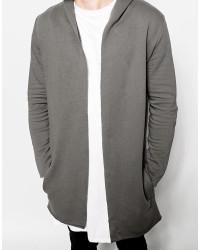}\label{Image}}
\subfigure[j)][]{\includegraphics[width=.375in,height=.375in]{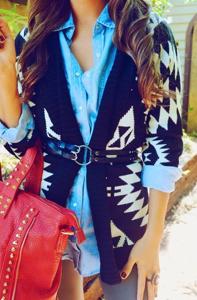}\label{Image}}\\[2pt]%

\subfigure{\includegraphics[width=.375in,height=.375in]{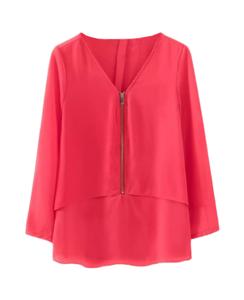}\label{Image}}
\subfigure{\includegraphics[width=.375in,height=.375in]{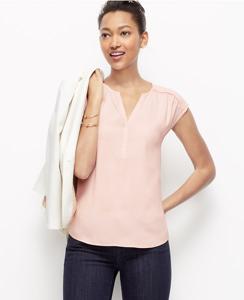}\label{Image}}
\subfigure{\includegraphics[width=.375in,height=.375in]{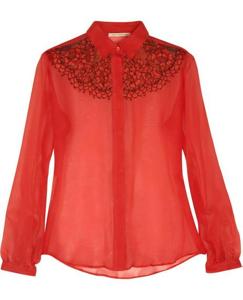}\label{Image}}
\subfigure{\includegraphics[width=.375in,height=.375in]{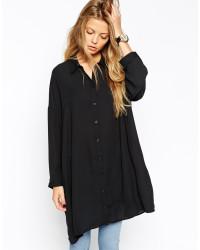}\label{Image}}
\subfigure{\includegraphics[width=.375in,height=.375in]{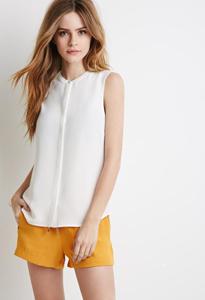}\label{Image}}
\subfigure{\includegraphics[width=.375in,height=.375in]{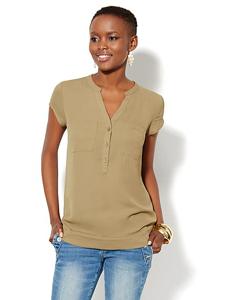}\label{Image}}
\subfigure{\includegraphics[width=.375in,height=.375in]{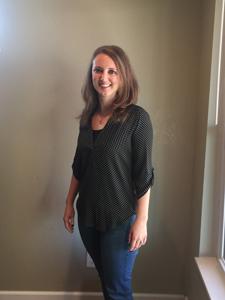}\label{Image}}
\subfigure{\includegraphics[width=.375in,height=.375in]{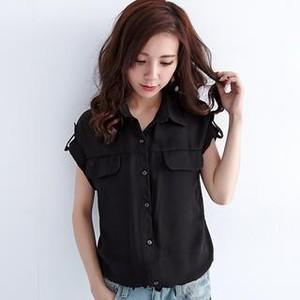}\label{Image}}
\subfigure{\includegraphics[width=.375in,height=.375in]{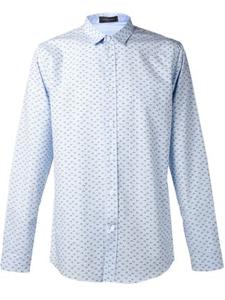}\label{Image}}
\subfigure{\includegraphics[width=.375in,height=.375in]{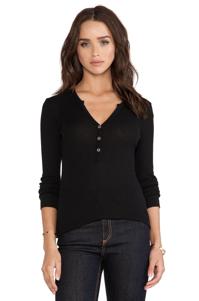}\label{Image}}\\[2pt]%

\subfigure{\includegraphics[width=.375in,height=.375in]{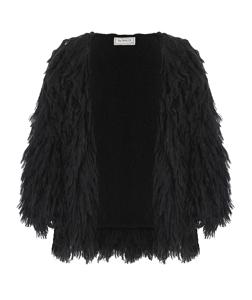}\label{Image}}
\subfigure{\includegraphics[width=.375in,height=.375in]{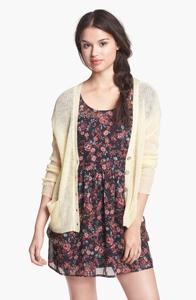}\label{Image}}
\subfigure{\includegraphics[width=.375in,height=.375in]{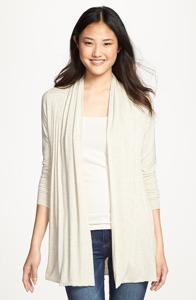}\label{Image}}
\subfigure{\includegraphics[width=.375in,height=.375in]{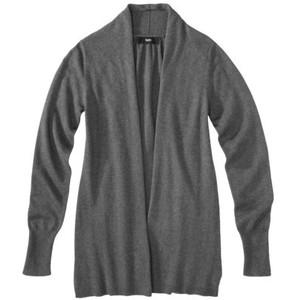}\label{Image}}
\subfigure{\includegraphics[width=.375in,height=.375in]{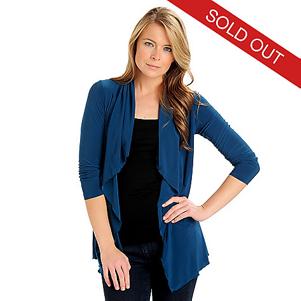}\label{Image}}
\subfigure{\includegraphics[width=.375in,height=.375in]{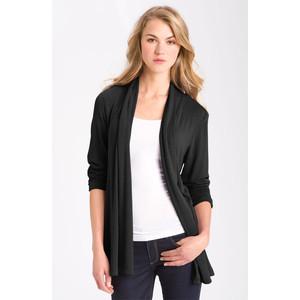}\label{Image}}
\subfigure{\includegraphics[width=.375in,height=.375in]{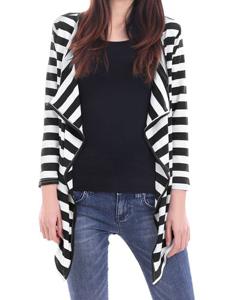}\label{Image}}
\subfigure{\includegraphics[width=.375in,height=.375in]{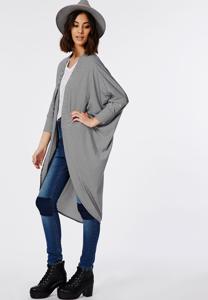}\label{Image}}
\subfigure{\includegraphics[width=.375in,height=.375in]{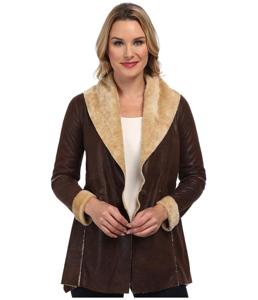}\label{Image}}
\subfigure{\includegraphics[width=.375in,height=.375in]{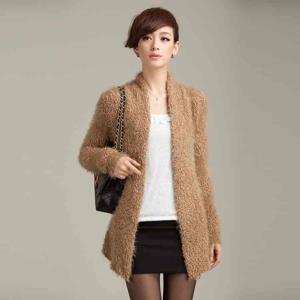}\label{Image}}\\[2pt]%

\subfigure{\includegraphics[width=.375in,height=.375in]{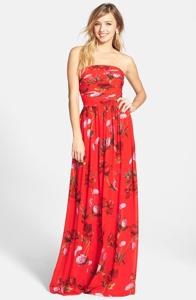}\label{Image}}
\subfigure{\includegraphics[width=.375in,height=.375in]{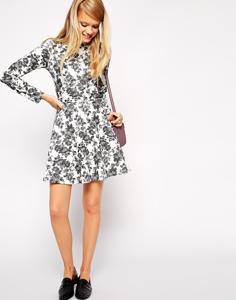}\label{Image}}
\subfigure{\includegraphics[width=.375in,height=.375in]{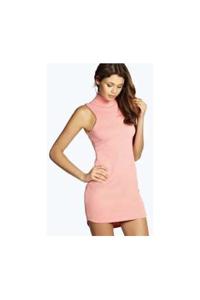}\label{Image}}
\subfigure{\includegraphics[width=.375in,height=.375in]{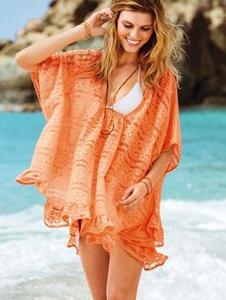}\label{Image}}
\subfigure{\includegraphics[width=.375in,height=.375in]{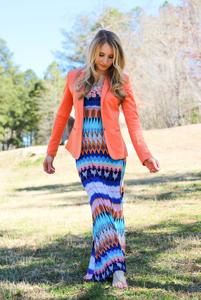}\label{Image}}
\subfigure{\includegraphics[width=.375in,height=.375in]{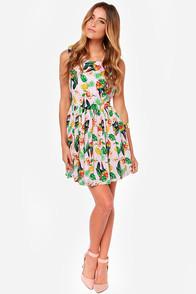}\label{Image}}
\subfigure{\includegraphics[width=.375in,height=.375in]{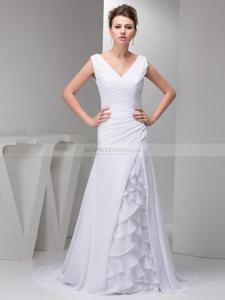}\label{Image}}
\subfigure{\includegraphics[width=.375in,height=.375in]{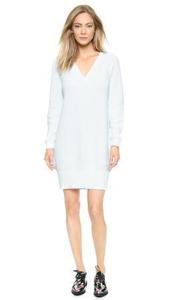}\label{Image}}
\subfigure{\includegraphics[width=.375in,height=.375in]{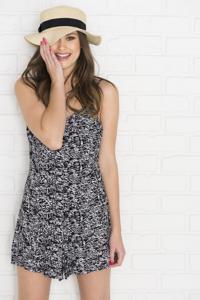}\label{Image}}
\subfigure{\includegraphics[width=.375in,height=.375in]{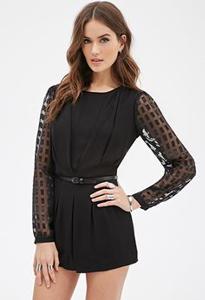}\label{Image}}\\[2pt]%

\subfigure{\includegraphics[width=.375in,height=.375in]{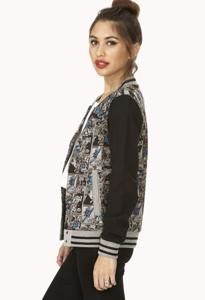}\label{Image}}
\subfigure{\includegraphics[width=.375in,height=.375in]{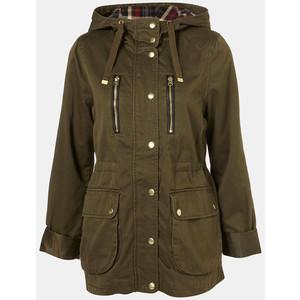}\label{Image}}
\subfigure{\includegraphics[width=.375in,height=.375in]{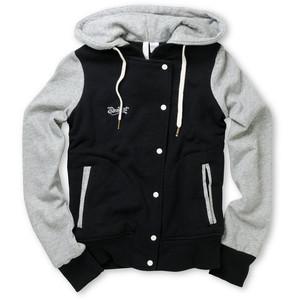}\label{Image}}
\subfigure{\includegraphics[width=.375in,height=.375in]{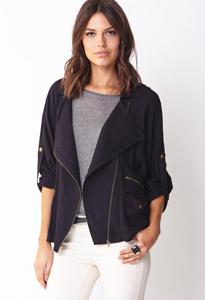}\label{Image}}
\subfigure{\includegraphics[width=.375in,height=.375in]{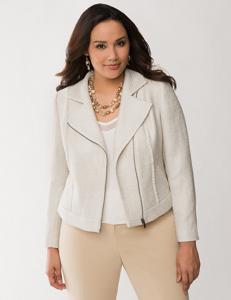}\label{Image}}
\subfigure{\includegraphics[width=.375in,height=.375in]{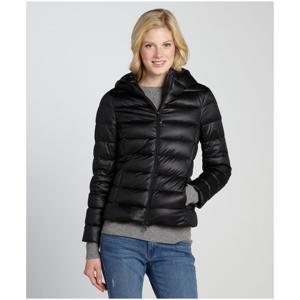}\label{Image}}
\subfigure{\includegraphics[width=.375in,height=.375in]{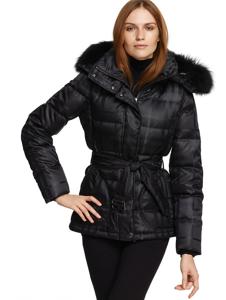}\label{Image}}
\subfigure{\includegraphics[width=.375in,height=.375in]{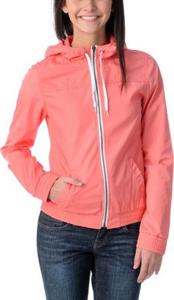}\label{Image}}
\subfigure{\includegraphics[width=.375in,height=.375in]{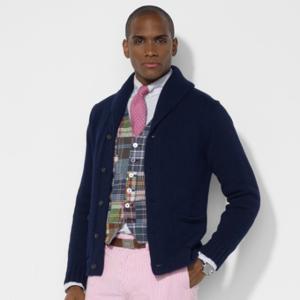}\label{Image}}
\subfigure{\includegraphics[width=.375in,height=.375in]{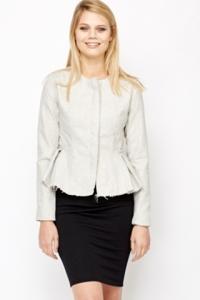}\label{Image}}\\[2pt]%

\subfigure{\includegraphics[width=.375in,height=.375in]{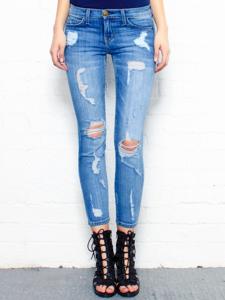}\label{Image}}
\subfigure{\includegraphics[width=.375in,height=.375in]{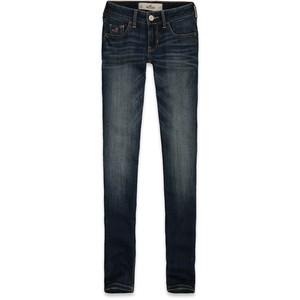}\label{Image}}
\subfigure{\includegraphics[width=.375in,height=.375in]{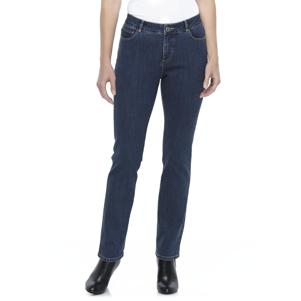}\label{Image}}
\subfigure{\includegraphics[width=.375in,height=.375in]{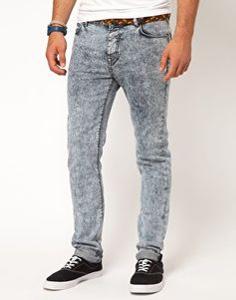}\label{Image}}
\subfigure{\includegraphics[width=.375in,height=.375in]{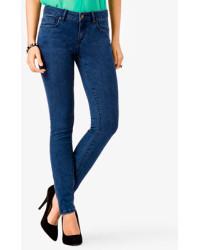}\label{Image}}
\subfigure{\includegraphics[width=.375in,height=.375in]{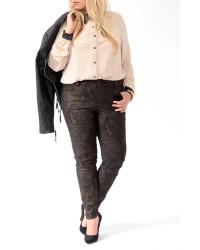}\label{Image}}
\subfigure{\includegraphics[width=.375in,height=.375in]{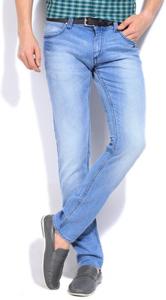}\label{Image}}
\subfigure{\includegraphics[width=.375in,height=.375in]{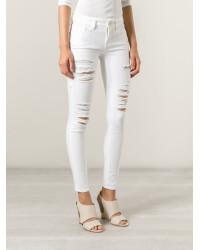}\label{Image}}
\subfigure{\includegraphics[width=.375in,height=.375in]{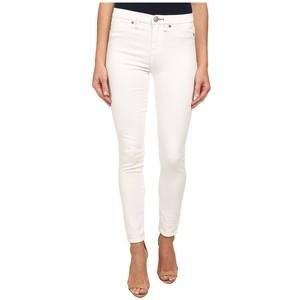}\label{Image}}
\subfigure{\includegraphics[width=.375in,height=.375in]{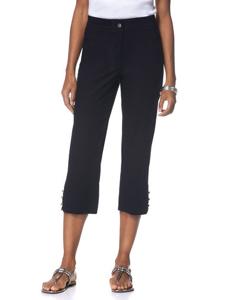}\label{Image}}\\[2pt]%

\subfigure{\includegraphics[width=.375in,height=.375in]{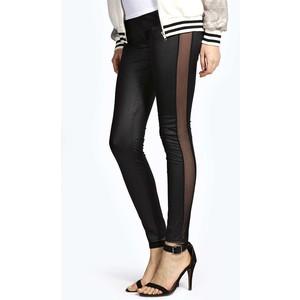}\label{Image}}
\subfigure{\includegraphics[width=.375in,height=.375in]{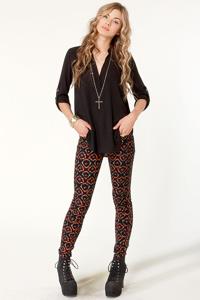}\label{Image}}
\subfigure{\includegraphics[width=.375in,height=.375in]{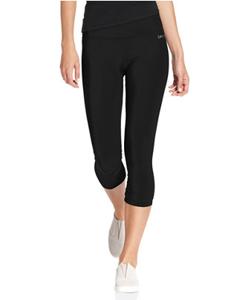}\label{Image}}
\subfigure{\includegraphics[width=.375in,height=.375in]{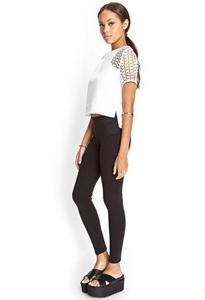}\label{Image}}
\subfigure{\includegraphics[width=.375in,height=.375in]{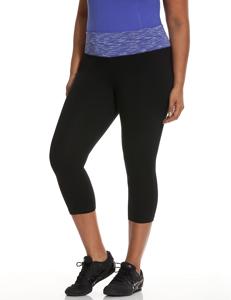}\label{Image}}
\subfigure{\includegraphics[width=.375in,height=.375in]{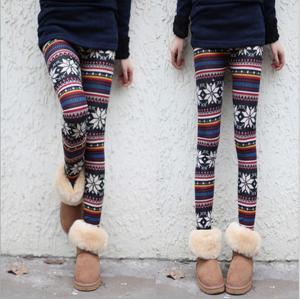}\label{Image}}
\subfigure{\includegraphics[width=.375in,height=.375in]{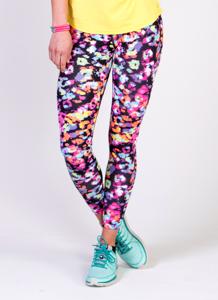}\label{Image}}
\subfigure{\includegraphics[width=.375in,height=.375in]{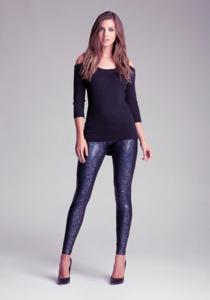}\label{Image}}
\subfigure{\includegraphics[width=.375in,height=.375in]{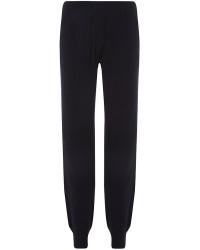}\label{Image}}
\subfigure{\includegraphics[width=.375in,height=.375in]{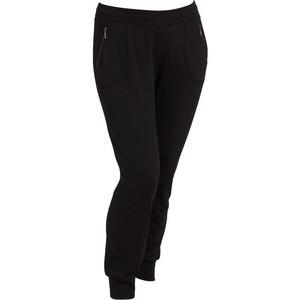}\label{Image}}\\[2pt]%

\subfigure{\includegraphics[width=.375in,height=.375in]{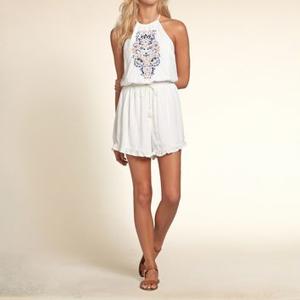}\label{Image}}
\subfigure{\includegraphics[width=.375in,height=.375in]{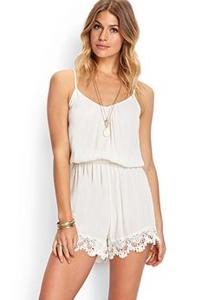}\label{Image}}
\subfigure{\includegraphics[width=.375in,height=.375in]{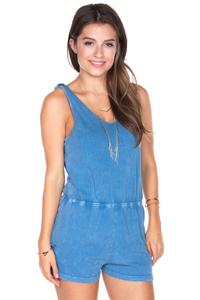}\label{Image}}
\subfigure{\includegraphics[width=.375in,height=.375in]{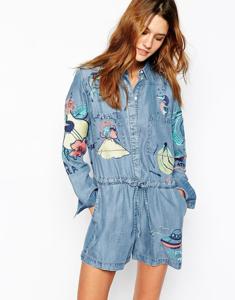}\label{Image}}
\subfigure{\includegraphics[width=.375in,height=.375in]{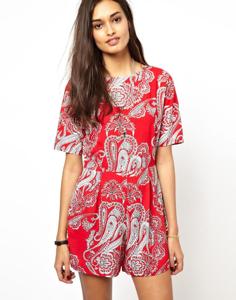}\label{Image}}
\subfigure{\includegraphics[width=.375in,height=.375in]{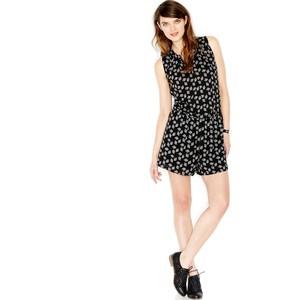}\label{Image}}
\subfigure{\includegraphics[width=.375in,height=.375in]{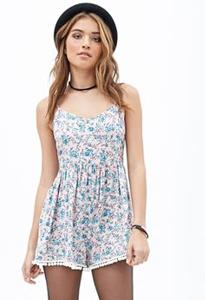}\label{Image}}
\subfigure{\includegraphics[width=.375in,height=.375in]{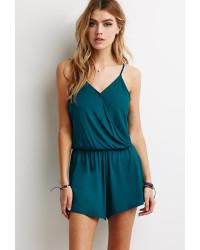}\label{Image}}
\subfigure{\includegraphics[width=.375in,height=.375in]{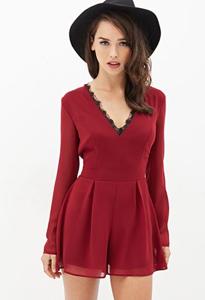}\label{Image}}
\subfigure{\includegraphics[width=.375in,height=.375in]{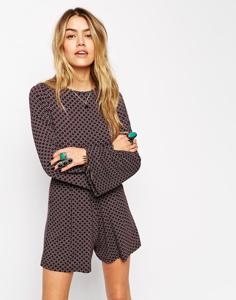}\label{Image}}\\[2pt]%

\subfigure{\includegraphics[width=.375in,height=.375in]{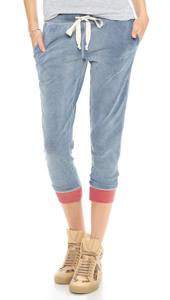}\label{Image}}
\subfigure{\includegraphics[width=.375in,height=.375in]{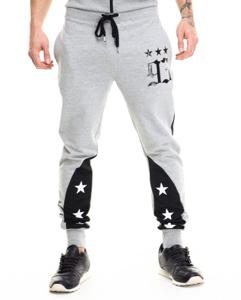}\label{Image}}
\subfigure{\includegraphics[width=.375in,height=.375in]{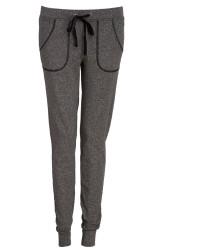}\label{Image}}
\subfigure{\includegraphics[width=.375in,height=.375in]{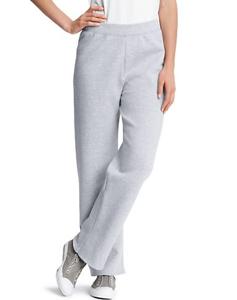}\label{Image}}
\subfigure{\includegraphics[width=.375in,height=.375in]{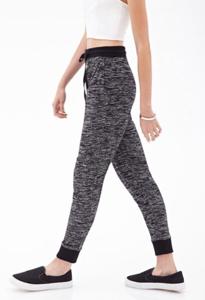}\label{Image}}
\subfigure{\includegraphics[width=.375in,height=.375in]{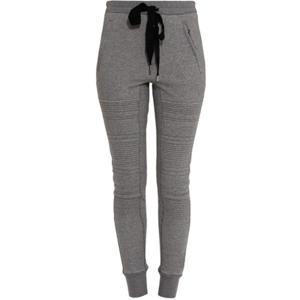}\label{Image}}
\subfigure{\includegraphics[width=.375in,height=.375in]{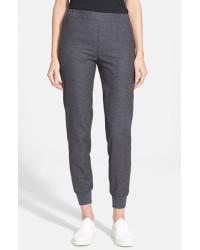}\label{Image}}
\subfigure{\includegraphics[width=.375in,height=.375in]{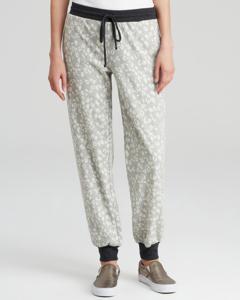}\label{Image}}
\subfigure{\includegraphics[width=.375in,height=.375in]{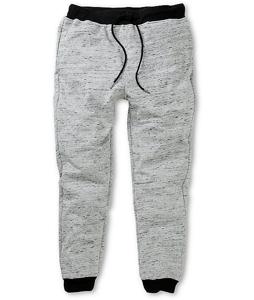}\label{Image}}
\subfigure{\includegraphics[width=.375in,height=.375in]{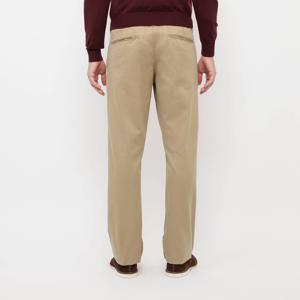}\label{Image}}\\[2pt]%
\subfigure{\includegraphics[width=.375in,height=.375in]{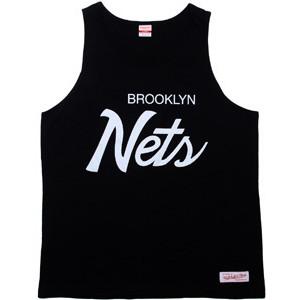}\label{Image}}
\subfigure{\includegraphics[width=.375in,height=.375in]{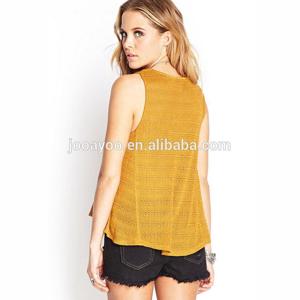}\label{Image}}
\subfigure{\includegraphics[width=.375in,height=.375in]{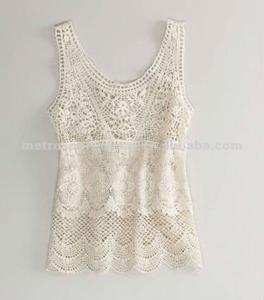}\label{Image}}
\subfigure{\includegraphics[width=.375in,height=.375in]{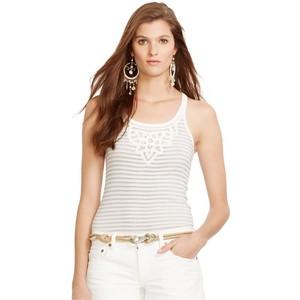}\label{Image}}
\subfigure{\includegraphics[width=.375in,height=.375in]{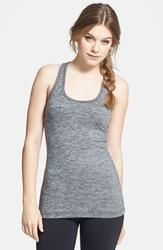}\label{Image}}
\subfigure{\includegraphics[width=.375in,height=.375in]{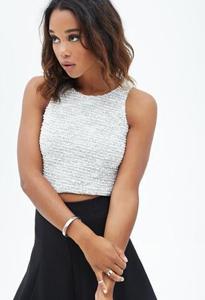}\label{Image}}
\subfigure{\includegraphics[width=.375in,height=.375in]{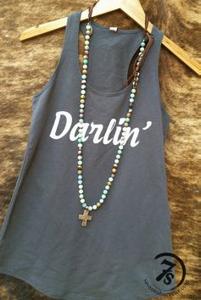}\label{Image}}
\subfigure{\includegraphics[width=.375in,height=.375in]{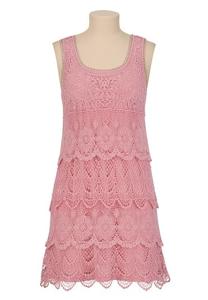}\label{Image}}
\subfigure{\includegraphics[width=.375in,height=.375in]{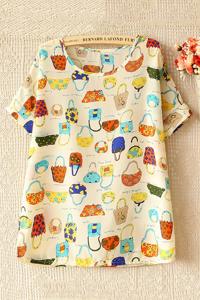}\label{Image}}
\subfigure{\includegraphics[width=.375in,height=.375in]{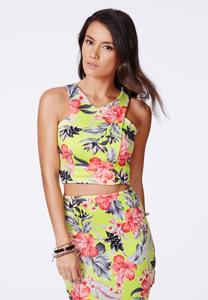}\label{Image}}\\[2pt]%
\subfigure{\includegraphics[width=.375in,height=.375in]{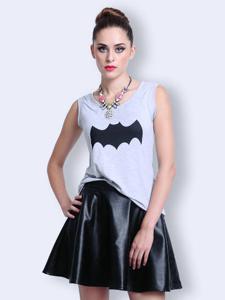}\label{Image}}
\subfigure{\includegraphics[width=.375in,height=.375in]{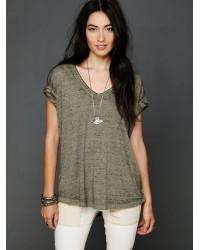}\label{Image}}
\subfigure{\includegraphics[width=.375in,height=.375in]{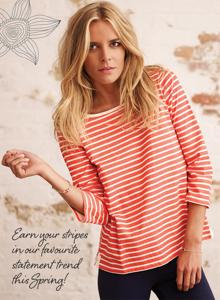}\label{Image}}
\subfigure{\includegraphics[width=.375in,height=.375in]{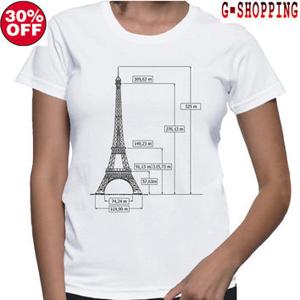}\label{Image}}
\subfigure{\includegraphics[width=.375in,height=.375in]{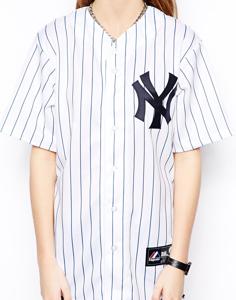}\label{Image}}
\subfigure{\includegraphics[width=.375in,height=.375in]{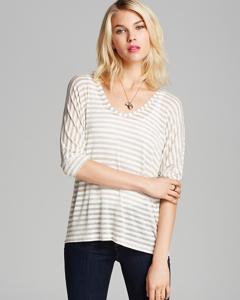}\label{Image}}
\subfigure{\includegraphics[width=.375in,height=.375in]{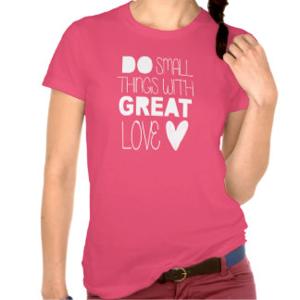}\label{Image}}
\subfigure{\includegraphics[width=.375in,height=.375in]{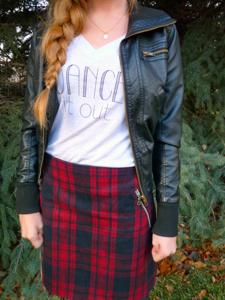}\label{Image}}
\subfigure{\includegraphics[width=.375in,height=.375in]{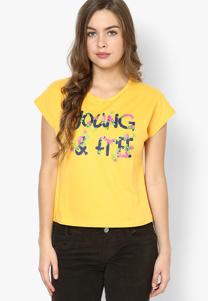}\label{Image}}
\subfigure{\includegraphics[width=.375in,height=.375in]{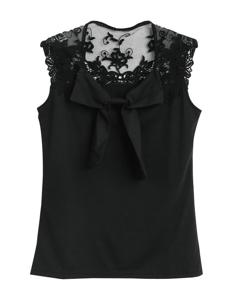}\label{Image}}\\[2pt]%

\caption{Results using our categorisation pipeline. (Rows 1-11) are Blazers, Blouses, Cardigans, Dresses, Jackets, Jeans, Leggings, Rompers, Sweatpants, Tanks and Tees respectively. Columns (a-h) are correctly classified images while columns (i-j) are incorrectly categorised images.(Best viewed in color)}
\label{fig:sop}
\vspace{2 mm}
\end{center}
\end{figure*}

\begin{figure}[!ht] 
\begin{center}
\subfigure[a)][]{\includegraphics[width=.6in,height=.6in]{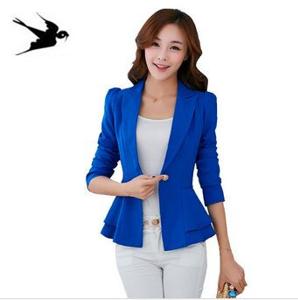}\label{Image}}
\subfigure[b)][]{\includegraphics[width=.6in,height=.6in]{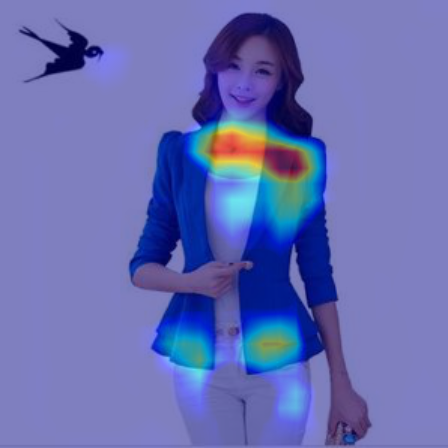}\label{Image}}
\subfigure[c)][]{\includegraphics[width=.6in,height=.6in]{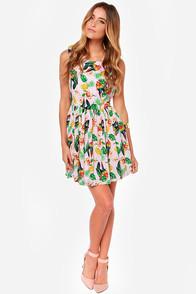}\label{Image}}
\subfigure[d)][]{\includegraphics[width=.6in,height=.6in]{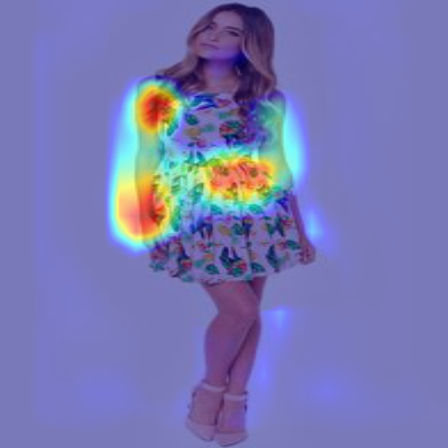}\label{Image}}
\subfigure[e)][]{\includegraphics[width=.6in,height=.6in]{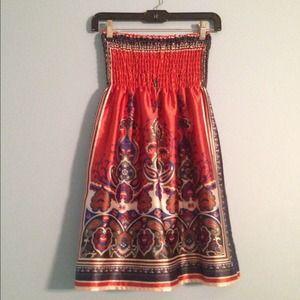}\label{Image}}
\subfigure[f)][]{\includegraphics[width=.6in,height=.6in]{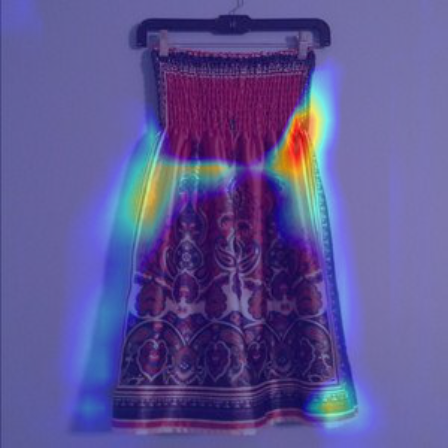}\label{Image}}

\subfigure{\includegraphics[width=.6in,height=.6in]{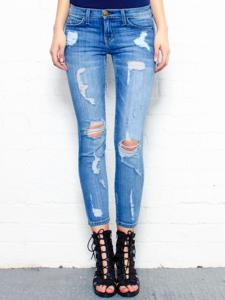}\label{Image}}
\subfigure{\includegraphics[width=.6in,height=.6in]{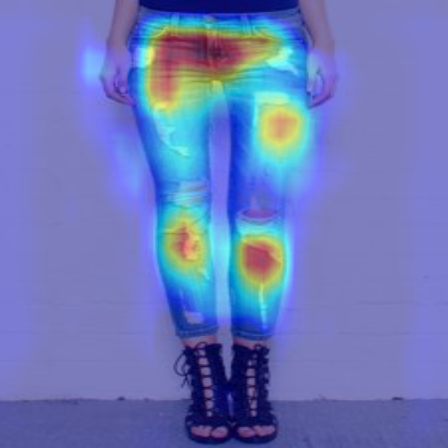}\label{Image}}
\subfigure{\includegraphics[width=.6in,height=.6in]{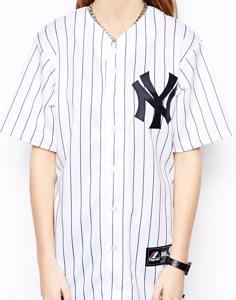}\label{Image}}
\subfigure{\includegraphics[width=.6in,height=.6in]{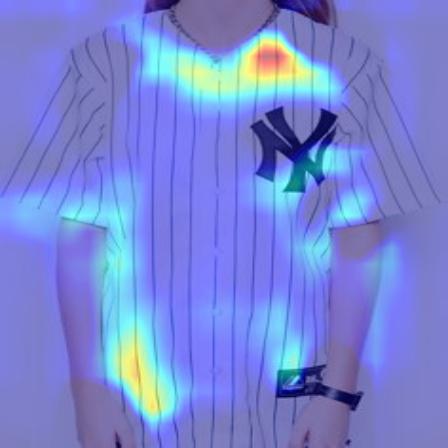}\label{Image}}
\subfigure{\includegraphics[width=.6in,height=.6in]{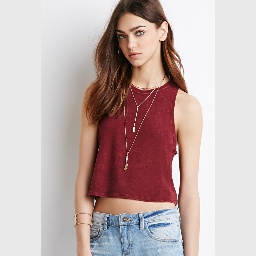}\label{Image}}
\subfigure{\includegraphics[width=.6in,height=.6in]{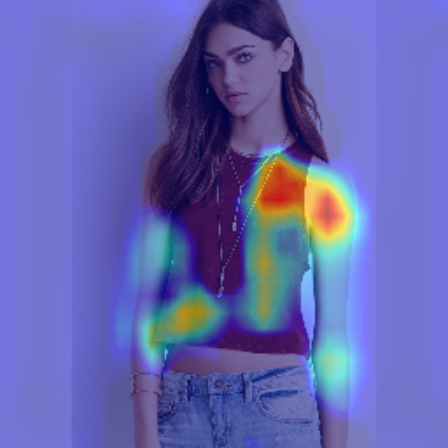}\label{Image}}

\subfigure{\includegraphics[width=.6in,height=.6in]{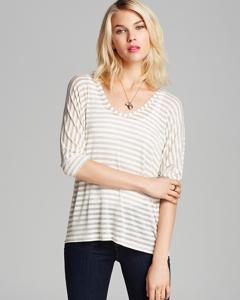}\label{Image}}
\subfigure{\includegraphics[width=.6in,height=.6in]{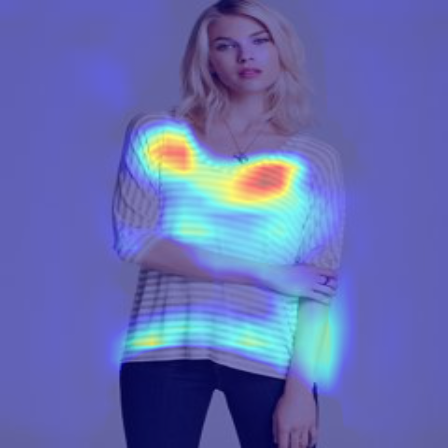}\label{Image}}
\subfigure{\includegraphics[width=.6in,height=.6in]{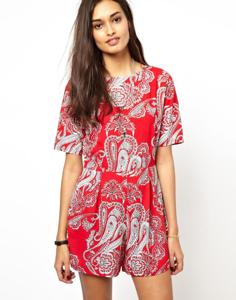}\label{Image}}
\subfigure{\includegraphics[width=.6in,height=.6in]{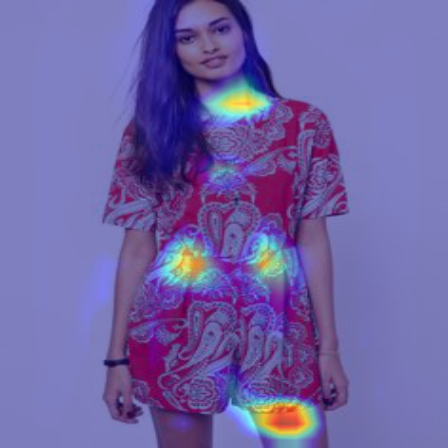}\label{Image}}
\subfigure{\includegraphics[width=.6in,height=.6in]{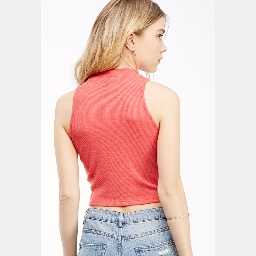}\label{Image}}
\subfigure{\includegraphics[width=.6in,height=.6in]{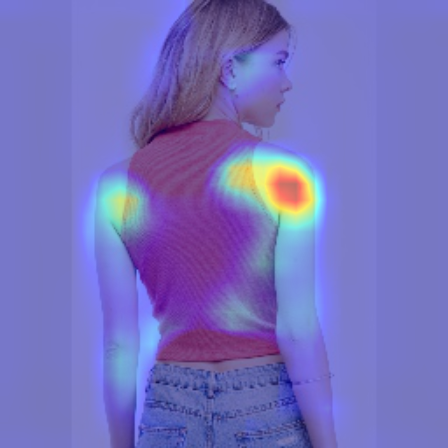}\label{Image}}

\caption{Columns (a, c and e) show \textit{actual images} while columns {b, d and f} depict \textit{Conv5-3 layer visualization
of the same using Grad-CAM \cite{selvaraju2017grad}} from proposed apparel categorization pipeline. (Best viewed in color)}

\label{fig:sop}
\vspace{-7 mm}
\end{center}
\end{figure}

\subsubsection{Fashion Apparel in-shop image retrieval}
In-shop image retrieval is difficult due to the task of detecting the same clothing item from different poses/ arrangement of the same. But still, our framework based on compact BCNN trained using triplet approach with hinge loss is able to gauge (dis)similarity between clothing apparel items without any additional requirement of finding bounding boxes, human joints~\cite{yang2011articulated}, poselets~\cite{bourdev2009poselets} or clothing landmarks~\cite{liu2016deepfashion}.
In fact, our framework is very close to the state of art FashionNet~\cite{liu2016deepfashion} which is trained using fashion landmarks, and clothing attribute information. Compact bilinear CNN trained without any bounding box annotation achieves a top- 20 accuracy of 76.26 while FashionNet achieves an overall accuracy of 76.4. Table II specifies a quantitative comparison of our in-shop retrieval framework while Figure 5 displays the qualitative results of the pipeline.

\begin{table}[t] 
\renewcommand{\arraystretch}{1}
\caption{Top-20 Accuracy for In-shop retrieval. We achieved these result without the dependence on part based models or rich annotations required for labelling datasets.}
\label{tab:Result2}
\centering
\begin{tabular}{|l|c|}
\hline \textbf{Method} & \textbf{Top-20}\\
\hline WTBI~\cite{chen2012describing} &     50.6\\
\hline DARN~\cite{huang2015cross} &    67.5\\
\hline FashionNet~\cite{liu2016deepfashion} &    76.4\\
\hline In-Shop Retrieval Framework & \textbf{76.26}\\
\hline
\end{tabular}
\end{table}

\begin{figure}[!ht] 
\begin{center}
\subfigure[a)][]{\includegraphics[width=.6in,height=.6in]{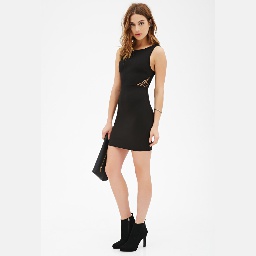}\label{Image}}
\subfigure[b)][]{\includegraphics[width=.6in,height=.6in]{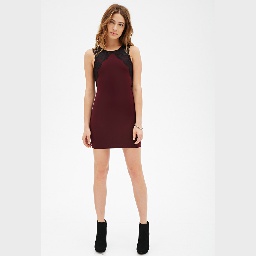}\label{Image}}
\subfigure[c)][]{\includegraphics[width=.6in,height=.6in]{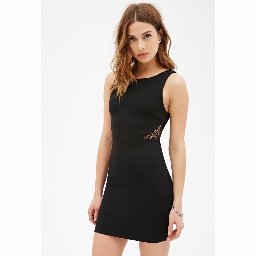}\label{Image}}
\subfigure[d)][]{\includegraphics[width=.6in,height=.6in]{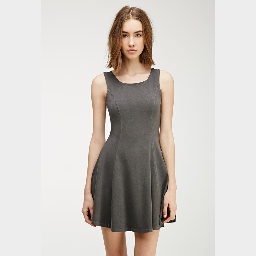}\label{Image}}
\subfigure[e)][]{\includegraphics[width=.6in,height=.6in]{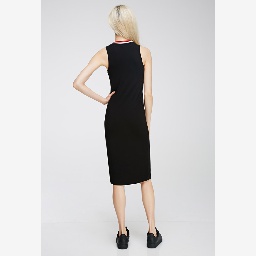}\label{Image}}
\subfigure[f)][]{\includegraphics[width=.6in,height=.6in]{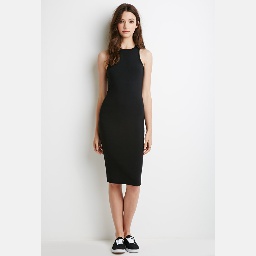}\label{Image}}\\[2pt]%

\subfigure{\includegraphics[width=.6in,height=.6in]{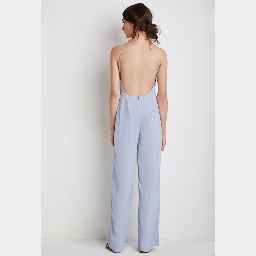}\label{Image}}
\subfigure{\includegraphics[width=.6in,height=.6in]{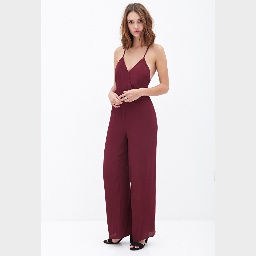}\label{Image}}
\subfigure{\includegraphics[width=.6in,height=.6in]{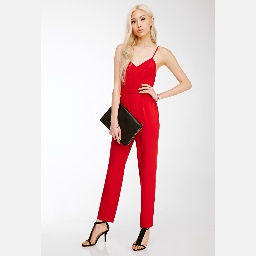}\label{Image}}
\subfigure{\includegraphics[width=.6in,height=.6in]{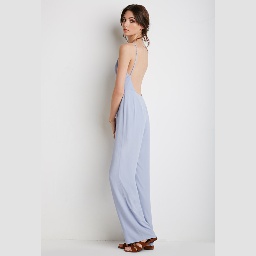}\label{Image}}
\subfigure{\includegraphics[width=.6in,height=.6in]{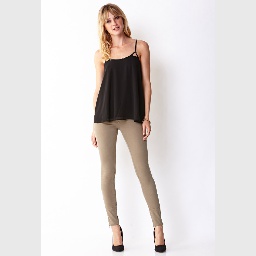}\label{Image}}
\subfigure{\includegraphics[width=.6in,height=.6in]{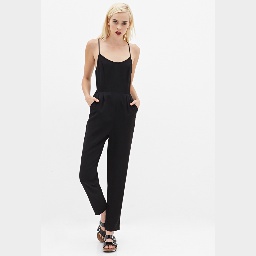}\label{Image}}\\[2pt]%

\subfigure{\includegraphics[width=.6in,height=.6in]{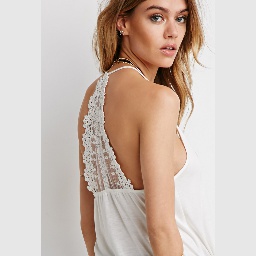}\label{Image}}
\subfigure{\includegraphics[width=.6in,height=.6in]{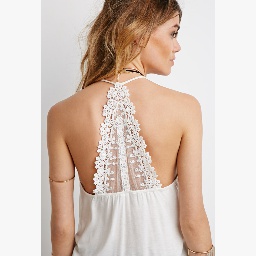}\label{Image}}
\subfigure{\includegraphics[width=.6in,height=.6in]{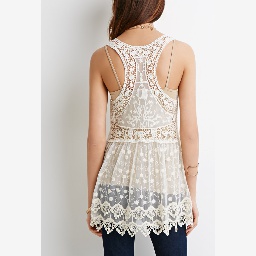}\label{Image}}
\subfigure{\includegraphics[width=.6in,height=.6in]{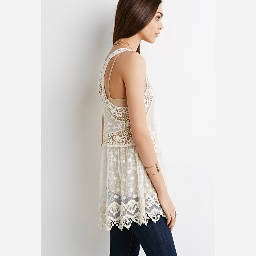}\label{Image}}
\subfigure{\includegraphics[width=.6in,height=.6in]{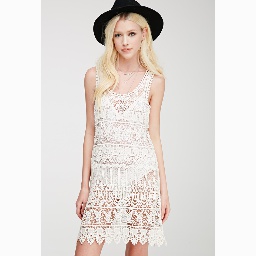}\label{Image}}
\subfigure{\includegraphics[width=.6in,height=.6in]{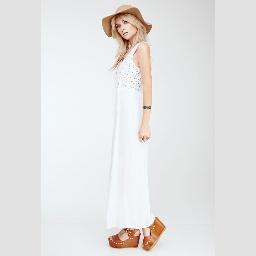}\label{Image}}\\[2pt]%

\subfigure{\includegraphics[width=.6in,height=.6in]{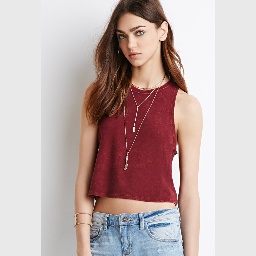}\label{Image}}
\subfigure{\includegraphics[width=.6in,height=.6in]{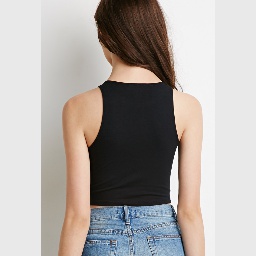}\label{Image}}
\subfigure{\includegraphics[width=.6in,height=.6in]{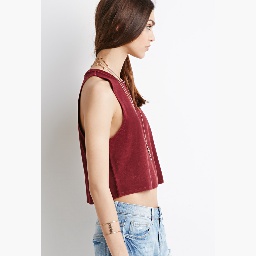}\label{Image}}
\subfigure{\includegraphics[width=.6in,height=.6in]{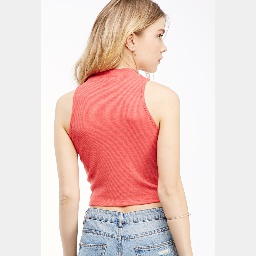}\label{Image}}
\subfigure{\includegraphics[width=.6in,height=.6in]{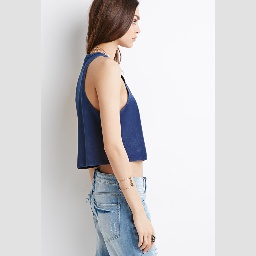}\label{Image}}
\subfigure{\includegraphics[width=.6in,height=.6in]{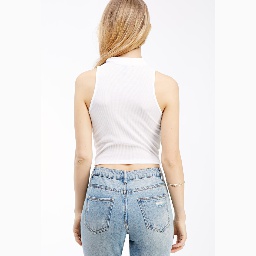}\label{Image}}\\[2pt]%

\caption{Columns (a-f) Query and Top-5 retrieved examples using in-shop retrieval pipeline. Column (a) Query Image, (b-f) Top 5 Images.(Best viewed in color)}
\label{fig:sop}
\vspace{2 mm}
\end{center}
\end{figure}

\subsubsection{Fashion Apparel cross-domain (street to shop) image retrieval}
Cross-domain image retrieval is even more difficult as compared to in-shop image retrieval because of the high variance between the images of two domains introduced by various factors such as bad lighting, less proportion of visible apparel item and clutter in the images because of surrounding items and background. Our framework outperforms WTBI and DARN and achieves a top-20 retrieval accuracy of 17.19 and is comparable to FashionNet~\cite{liu2016deepfashion} which is trained using bounding box and fashion landmarks jointly with clothing attributes.
Table III specifies a quantitative comparison of our cross-domain retrieval framework while Figure 6 displays the qualitative results of the pipeline.

\begin{table}[t] 
\renewcommand{\arraystretch}{1}
\caption{Top-20 Accuracy for Cross-domain retrieval. We achieved these result without the dependence on Part based models or rich annotations required for labelling datasets.}
\label{tab:Result3}
\centering
\begin{tabular}{|l|c|c|}
\hline \textbf{Method} & \textbf{Top-20 Retrieval Accuracy}\\
\hline WTBI~\cite{chen2012describing} &     6.3\\
\hline DARN~\cite{huang2015cross} &    11.1\\
\hline FashionNet~\cite{liu2016deepfashion} &    18.8\\
\hline Cross-Domain Retrieval Framework & \textbf{17.49}\\
\hline
\end{tabular}
\end{table}

\begin{figure}[!ht] 
\begin{center}
\subfigure[a)][]{\includegraphics[width=.6in,height=.6in]{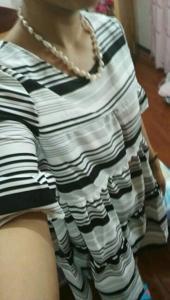}\label{Image}}
\subfigure[b)][]{\includegraphics[width=.6in,height=.6in]{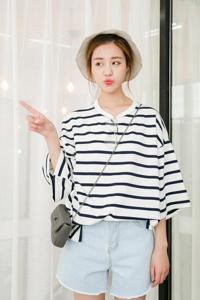}\label{Image}}
\subfigure[c)][]{\includegraphics[width=.6in,height=.6in]{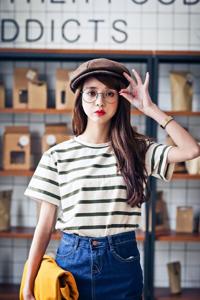}\label{Image}}
\subfigure[d)][]{\includegraphics[width=.6in,height=.6in]{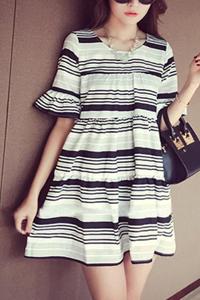}\label{Image}}
\subfigure[e)][]{\includegraphics[width=.6in,height=.6in]{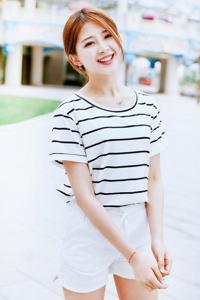}\label{Image}}
\subfigure[f)][]{\includegraphics[width=.6in,height=.6in]{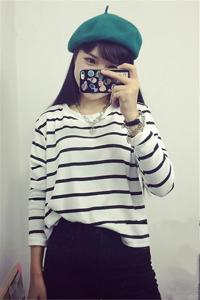}\label{Image}}\\[2pt]%

\subfigure{\includegraphics[width=.6in,height=.6in]{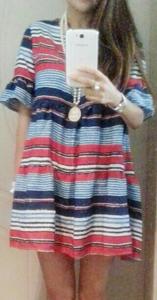}\label{Image}}
\subfigure{\includegraphics[width=.6in,height=.6in]{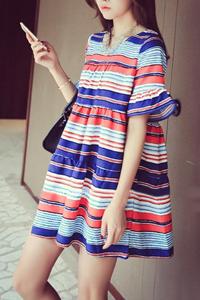}\label{Image}}
\subfigure{\includegraphics[width=.6in,height=.6in]{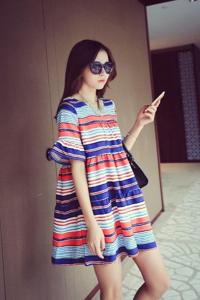}\label{Image}}
\subfigure{\includegraphics[width=.6in,height=.6in]{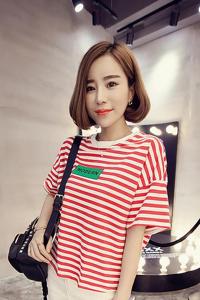}\label{Image}}
\subfigure{\includegraphics[width=.6in,height=.6in]{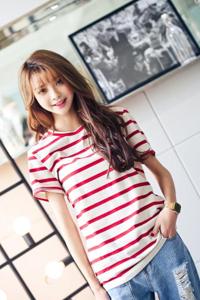}\label{Image}}
\subfigure{\includegraphics[width=.6in,height=.6in]{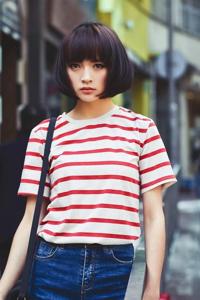}\label{Image}}\\[2pt]%

\subfigure{\includegraphics[width=.6in,height=.6in]{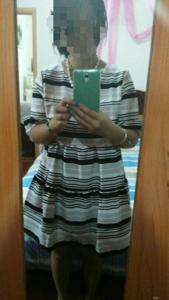}\label{Image}}
\subfigure{\includegraphics[width=.6in,height=.6in]{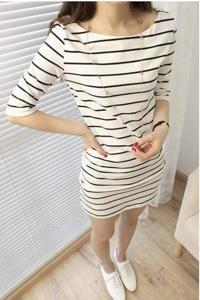}\label{Image}}
\subfigure{\includegraphics[width=.6in,height=.6in]{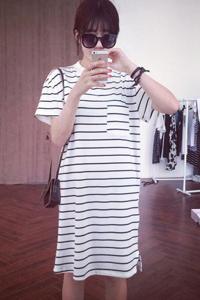}\label{Image}}
\subfigure{\includegraphics[width=.6in,height=.6in]{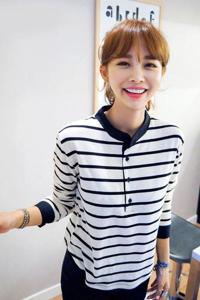}\label{Image}}
\subfigure{\includegraphics[width=.6in,height=.6in]{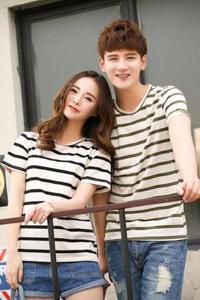}\label{Image}}
\subfigure{\includegraphics[width=.6in,height=.6in]{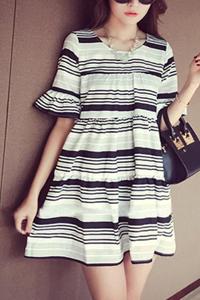}\label{Image}}\\[2pt]%


\subfigure{\includegraphics[width=.6in,height=.6in]{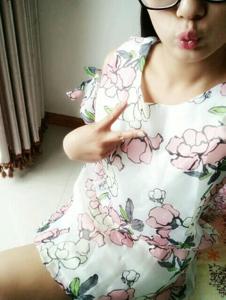}\label{Image}}
\subfigure{\includegraphics[width=.6in,height=.6in]{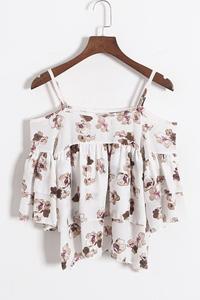}\label{Image}}
\subfigure{\includegraphics[width=.6in,height=.6in]{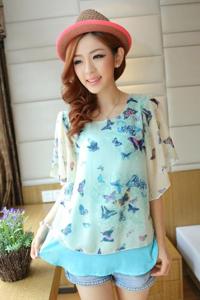}\label{Image}}
\subfigure{\includegraphics[width=.6in,height=.6in]{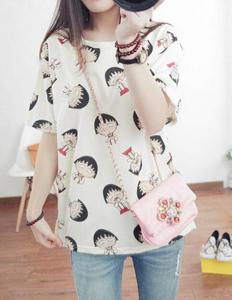}\label{Image}}
\subfigure{\includegraphics[width=.6in,height=.6in]{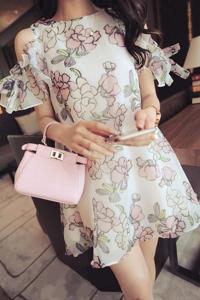}\label{Image}}
\subfigure{\includegraphics[width=.6in,height=.6in]{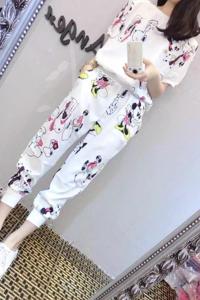}\label{Image}}\\[2pt]%

\caption{Columns (a-f) shows \textit{query image} and \textit{Top-5} retrieved examples using our proposed cross-domain retrieval pipeline. Column a) is \textit{ query image} and column(b-f) are Top 5 retrieved images.(Best viewed in color)}
\label{fig:sop}
\vspace{-7 mm}
\end{center}
\end{figure}


\section{DISCUSSION AND CONCLUSION}

In this paper, we proposed a framework for apparel categorization, and in-shop and cross-domain apparel retrieval problems. We have shown that compact bilinear CNNs can be duly utilized for fine-grained clothing categorization and similar apparel retrieval. Our experimental results show that our framework outperforms and is comparable with state-of-the-art methods without the dependence to obtain bounding boxes around the apparel item.

Further, it eliminates the requirement to train human joints, pose-lets or clothing landmark detectors to combat high intra-class variance in clothing datasets due to different poses, styles and non-rigid deformations in the apparel items, which has been a major prerequisite to classification in previous fashion classification or retrieval research. As a future work in this domain, it would be interesting to see if we can infuse the concept of attention with bilinear models to push the state of art further in this domain.


\bibliographystyle{spmpsci}
\bibliography{References}
\end{document}